\documentclass[letterpaper]{article} % DO NOT CHANGE THIS
\usepackage{aaai2026}  % DO NOT CHANGE THIS
\usepackage{times}  % DO NOT CHANGE THIS
\usepackage{helvet}  % DO NOT CHANGE THIS
\usepackage{courier}  % DO NOT CHANGE THIS
\usepackage[hyphens]{url}  % DO NOT CHANGE THIS
\usepackage{graphicx} % DO NOT CHANGE THIS
\usepackage{natbib}  % DO NOT CHANGE THIS AND DO NOT ADD ANY OPTIONS TO IT
\usepackage{caption} % DO NOT CHANGE THIS AND DO NOT ADD ANY OPTIONS TO IT
\usepackage[utf8]{inputenc} % allow utf-8 input
\usepackage[T1]{fontenc}    % use 8-bit T1 fonts
\usepackage{booktabs}       % professional-quality tables
\usepackage{amsfonts}       % blackboard math symbols
\usepackage{nicefrac}       % compact symbols for 1/2, etc.
\usepackage{microtype}      % microtypography
\usepackage{array}
\usepackage{pifont}
\usepackage{diagbox}
\usepackage[ruled]{algorithm2e}
\usepackage{enumitem}
\usepackage{amsmath}
\usepackage{multirow}
\usepackage{colortbl}
\usepackage{listings}
\newcommand{\cmark}{\ding{51}}
\newcommand{\xmark}{\ding{55}}

\nocopyright

\title{Interest Entanglement: The Hidden Barrier to Blind Super-Resolution Optimization}
\author{
    Junxiong Lin\textsuperscript{\rm 1},
    Xinji Mai\textsuperscript{\rm 1},
    Qianyu Guo\textsuperscript{\rm 4},
    Haoran Wang\textsuperscript{\rm 1},\\
    Zeng Tao\textsuperscript{\rm 1},
    Xuan Tong\textsuperscript{\rm 1},
    Ivy Pan\textsuperscript{\rm 3},
    Wenqiang Zhang\textsuperscript{\rm 1,\rm 2}
}
\affiliations{
    \textsuperscript{\rm 1}College of Intelligent Robotics and Advanced Manufacturing, Fudan University\\
    \textsuperscript{\rm 2}College of Computer Science and Artificial Intelligence, Fudan University\\
    \textsuperscript{\rm 3}The University of Hong Kong\\
    \textsuperscript{\rm 4}Shanghai Institute of Virology, Shanghai Jiao Tong University School of Medicine
}

\begin{document}

\maketitle

% ---- figure ----  
\begin{figure*}[t]
\centering
 \includegraphics[width=\textwidth]{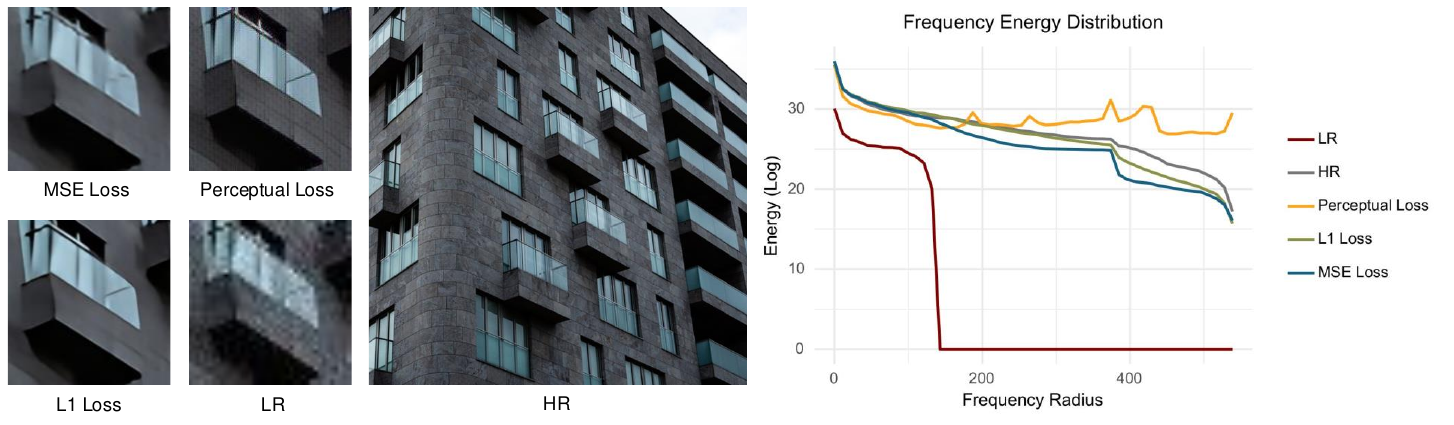}
 \caption{Frequency domain energy distribution of High-Resolution (HR) images, Low-Resolution (LR) images, and Super-Resolution images trained with different losses. Regression loss and perception loss focus on different aspects in the frequency domain. In Super-Resolution images, MSE loss and L1 loss result in localized over-smoothing, while Perceptual loss introduces undesirable artifacts.}
   \label{fig:motivation}
\end{figure*}
% ---- ---- ---- ---- ----

\begin{abstract}
Fidelity and perceptual quality are two inherently competing and conflicting objectives in the image super-resolution (SR) task. Different loss functions focus on these objectives to varying extents. Regression losses enhance the model’s fidelity but lack sufficient attention to high-frequency details, resulting in a loss of fine details. In contrast, perception losses improve the model’s visual quality but may introduce undesirable artifacts. Balancing these two optimization goals can be viewed as a Multi-Objective Optimization problem. Existing methods are limited to cautiously adjusting weight parameters between these losses, overlooking the underlying Interest Entanglement problem. To address this problem, we explore the inherent frequency-domain conflict between the regression objective and the perceptual objective, and analyze the causes of Interest Entanglement in SR tasks. According to our findings, we propose the Shared-Feature-Representation based Super-Resolution framework (SFR), which decouples the learning process of different optimization objectives, allowing the model to explore a common optimization direction for both goals and achieve an effective balance between them. To better leverage shared features, we also proposed the InfoSqueeze module, which filters redundant information through a dimensionality reduction and expansion process, effectively transforming features into a consistent space. Quantitative and qualitative experiments across five representative datasets affirm the superiority of SFR. 
\end{abstract}

\section{Introduction}
\label{sec:intro}

Image Super-Resolution, a highly regarded task within the foundational domain of computer vision, seeks to reconstruct HR images from LR observations \cite{xia2023meta,sun2023coser,wei2020component}. By augmenting pixel count, it aims to enhance image quality, thereby better serving downstream tasks.

Traditional research primarily evaluated model performance using pixel-based metrics such as PSNR and SSIM, leading to the adoption of regression losses like L1 loss and MSE loss for training \cite{vu2018perception}. However, these regression losses often result in the loss of high-frequency details, causing localized over-smoothing \cite{sun2024perception}, as illustrated in Figure \ref{fig:motivation}. Recently, some works have shifted their focus to the visual quality of SR models, using perceptual metrics like LPIPS for evaluation. To perform better on perceptual metrics, models often incorporate perception losses during training, enabling better restoration of high-frequency details. However, this approach frequently introduces artifacts or distorted textures in the SR results.

Given the inherent conflict between regression and perception losses, researchers must carefully balance the model’s performance across these two aspects during training. A common approach is to pre-train the model on regression loss to establish basic pixel-level SR capabilities, followed by fine-tuning on perception loss with a smaller learning rate and fewer epochs to achieve a compromise in visual perceptual quality \cite{bell2019blind,Zamir2021Restormer,chen2023hat,lin2024suppressing,lin2024adaptive,wang2023exploiting}. However, this training process is neither systematic nor efficient, often making it challenging to attain an optimal balance between the two objectives. Some studies have framed this issue as a Multi-Objective Optimization problem, employing parameter searches to balance the two types of loss. Yet, a straightforward linear combination of these losses overlooks the inherent ``Interest Entanglement" problem, thereby limiting the model’s potential performance \cite{pan2024ads,sener2018multi,mai2026agentic,mai2025cues}.

Interest Entanglement refers to the phenomenon in multi-objective learning where conflicting or inconsistent preferences among different task objectives lead to confusion or conflict in the learned feature representations \cite{su2024stem,shen2022hierarchically}. In Figure \ref{fig:motivation}, we analyze the distribution of energy across different frequencies in the frequency domain for various images. As seen in Figure \ref{fig:motivation}, LR images lose substantial high-frequency information compared to HR images, and different loss functions exhibit varying levels of focus on high-frequency information. Regression losses (L1 Loss, MSE Loss) evidently pay less attention to high-frequency details than perceptual-based losses, though perception loss (Perceptual Loss) do not match the fidelity of regression losses in reconstructing HR images. In the later section, we will further analyze and illustrate the root causes of Interest Entanglement in SR tasks.

In the blind image SR tasks, the degradation process is unknown, often comprising complex degradations such as blur, noise, resizing, and JPEG compression \cite{xia2022knowledge,wang2021unsupervised}. Consequently, reconstructing HR images while balancing regression and perception losses presents a substantial challenge. To address the Interest Entanglement issue in blind SR tasks, we propose a Shared-Feature-Representation based framework (SFR). In SFR, distinct modules are designed to learn different reconstruction objectives, while a shared feature representation module enables the model to capture common features across varied optimization targets. 

In summary, our primary contributions are as follows:
\begin{itemize}
% [leftmargin=*, noitemsep, topsep=0pt, partopsep=0pt]
    \item We are the first to identify and analyze the Interest Entanglement problem between regression and perceptual objectives in image SR, a conflict whose underlying causes have not been explored.
    
    \item Based on our findings, we propose the Shared-Feature-Representation based Super-Resolution framework (SFR). Additionally, we introduced the InfoSqueeze module to better transform Share-Feature within SFR.
    
    \item Extensive experiments across multiple representative datasets demonstrate the efficacy of SFR. Comprehensive qualitative, quantitative, and ablation studies underscore the effectiveness of our proposed approach.
\end{itemize}

\section{Related Work}
\label{sec:related}
\subsection{Blind Image Super-Resolution}
Distinct from classic Single Image Super-Resolution tasks, the objective of blind SRis to reconstruct high-resolution images from low-resolution inputs under an unknown degradation process \cite{wang2021real,luo2020unfolding,zhang2021designing}. Some approaches explicitly model the degradation process via blur kernels \cite{gu2019blind,yue2022blind,wang2023exploiting}. To achieve more precise blur kernel estimation. IKC \cite{gu2019blind} proposes an iterative estimation framework coupled with a correction function. Leveraging internal cross-scale recurrence, KernelGAN \cite{bell2019blind} maximizes patch recurrence within a single image, framing it as a data distribution learning problem and training a Generative Adversarial Network \cite{creswell2018generative} between patches. MANet \cite{liang2021mutual} employs a spatially variant blur kernel estimation approach, utilizing a network with an optimally sized receptive field to estimate the blur kernel. However, these approaches generally fail to address degradation patterns beyond blurring. Alternatively, some methods implicitly model the degradation process, with DASR \cite{wang2021unsupervised} and KDSR \cite{xia2022knowledge} utilizing contrastive learning and knowledge distillation, respectively, to capture characteristics of image degradation.

Some methods also directly learn degradation patterns from training data in the form of high-level semantics. SwinIR \cite{liang2021swinir}, by adopting the Swin Transformer for image restoration tasks, has achieved breakthrough performance. Works such as Restormer \cite{Zamir2021Restormer}, HAT \cite{chen2023activating}, and DAT \cite{chen2023dual} further demonstrate the potential of Vision Transformers in low-level vision tasks. Additionally, some researchers have focused on diffusion models \cite{rombach2021highresolution, wang2022zero, yue2024resshift, wang2023exploiting}, which transform complex and unstable generative processes into several independent and stable reverse processes through Markov chain modeling.

Given the more pronounced degradation in blind SR tasks, achieving an optimal balance between regression loss and perception loss presents a heightened challenge \cite{ai2023sosr,zhang2024real,vu2018perception}.
% %-------------------------------------------------------------------------
\subsection{Multi-objective Optimization in Image Super-Resolution}

Multi-Objective Optimization (MOO) refers to scenarios where multiple objective functions need to be optimized simultaneously within a single problem \cite{zitzler2001spea2,nebro2009smpso}. Typically, these objectives conflict with each other, meaning that improving one objective might lead to a decrease in performance for another \cite{sener2018multi}. Therefore, the key in MOO is to find a balance point among all objectives, achieving an optimal compromise. Classic MOO algorithms aim to identify a Pareto optimal solution set, in which no solution can be improved in one objective without compromising another \cite{jones1998efficient,emmerich2006single,tian2021evolutionary}. With the advancement of deep learning, many efforts have emerged to leverage neural networks to address this challenge \cite{su2024stem,tang2024multi}.

In the field of image Super-Resolution, fidelity and perceptual quality can be viewed as distinct optimization objectives \cite{vu2018perception}. Nearly all SRmethods incorporate pixel-based loss functions (such as L1 Loss or MSE Loss) during training. To mitigate the perceptual quality degradation associated with pixel-based losses, researchers often fine-tune the model with perceptual loss \cite{johnson2016perceptual} or GAN loss \cite{creswell2018generative} toward the end of training \cite{chen2023hat,xia2022knowledge,chen2023dual}. However, this approach heavily relies on manual tuning, leading to instability in training and difficulty in achieving optimal results.

Recently, some researchers have approached the balance between perceptual quality and distortion from an MOO perspective \cite{Park_Son_Cho_Hong_Lee_2018,Sajjadi_Scholkopf_Hirsch_2017}. ESRGAN \cite{Wang_Yu_Wu_Gu_Liu_Dong_Qiao_Loy_2019}, for instance, uses network parameter interpolation to balance these two objectives. \cite{sun2024perception} manually assigns different weights to various losses, while \cite{zhu2024perceptual} dynamically adjusts these weights during model training, removing the need for hyperparameter tuning. TGSR \cite{zhang2024real} reduces the negative impact of task competition by grouping unsatisfactory degradation tasks. However, these approaches remain focused on tuning the weights of different losses, overlooking the potential issue of inconsistent optimization directions among losses.

\section{Method}
In this section, we will analyze the root causes of the Interest Entanglement problem in SR tasks from a frequency domain perspective. We aim to explore why different loss functions exhibit distinct optimization focuses and, based on this understanding, propose our solution.
% ---- figure ----  
\begin{figure*}[t]
  \centering
  % \fbox{\rule{0pt}{2in} \rule{0.97\linewidth}{0pt}}
    \includegraphics[width=0.9\linewidth]{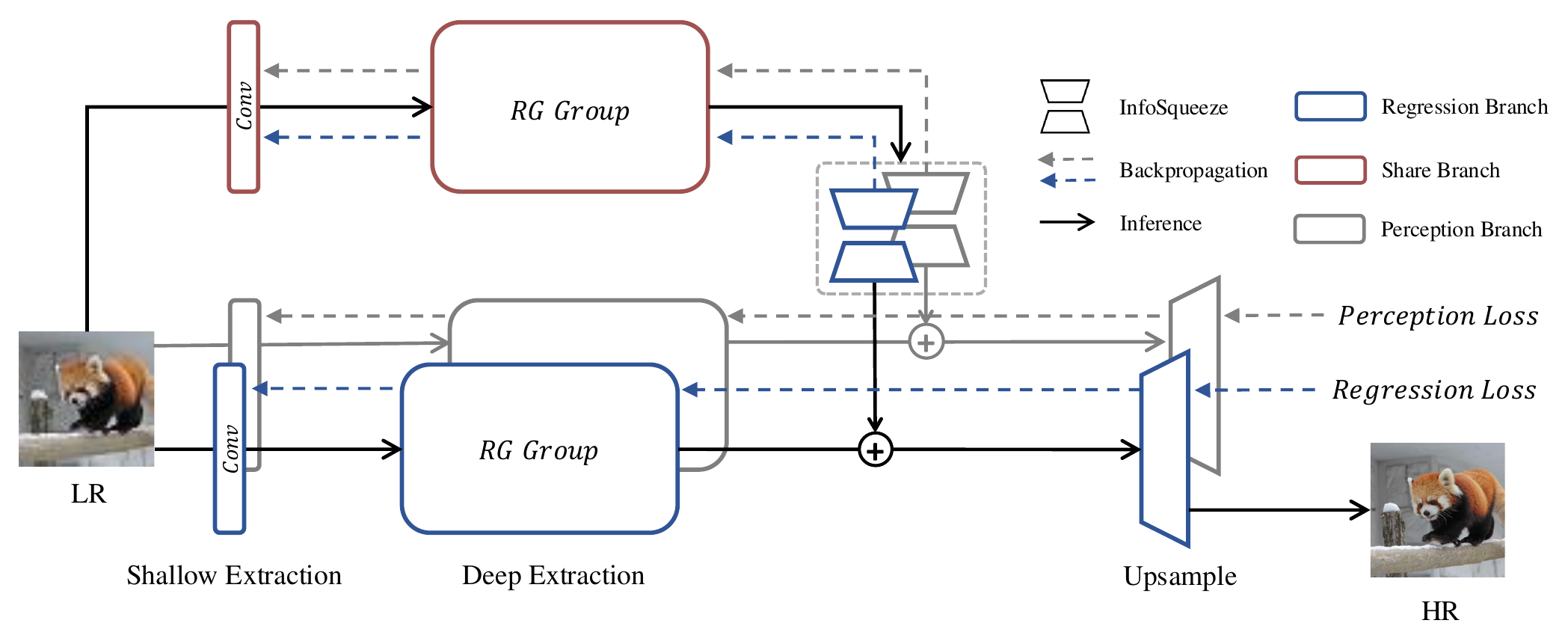}
   \caption{The Shared-Feature-Representation based Super-Resolution framework (SFR). SFR consists of three branches: the Regression Branch, Perception Branch, and Share Branch. Each branch mainly consists of Shallow Extraction, Deep Extraction, and Upsample modules. InfoSqueeze performs feature transformation on the feature extracted by the Share Branch.}
   \label{fig:framework}
\end{figure*}
% ---- ---- ---- ---- ----

\subsection{Interest Entanglement in SR: Frequency Domain Perspective}
Some researchers have recognized the task competition between different losses, yet the underlying causes remain unexplored. Specifically, regression losses and perception losses inherently optimize image quality in terms of fidelity and perceptual appeal, respectively. In this section, we select L1 Loss and Perceptual Loss \cite{johnson2016perceptual} as representatives of these two loss categories and analyze the root causes of this phenomenon from a frequency domain perspective.

\noindent \textbf{L1 Loss.} In the pixel space, the L1 loss between the super-resolved image $\hat{f}$ and the target image $f$ is defined as:
\begin{equation}
    L_{\text{L}1}=\frac1N\sum_{i=1}^N|f_i(x,y)-\widehat{f}_i(x,y)|
\end{equation}
\noindent Applying the Fourier transform \cite{sneddon1995fourier}, we obtain:
\begin{align}
    F_i(u,v) &= \sum_{x=0}^{M-1}\sum_{y=0}^{N-1} f_i(x,y)e^{-j2\pi\left(\frac{ux}{M} + \frac{vy}{N}\right)} \\
    \widehat{F}_i(u,v) &= \sum_{x=0}^{M-1}\sum_{y=0}^{N-1} \widehat{f}_i(x,y)e^{-j2\pi\left(\frac{ux}{M} + \frac{vy}{N}\right)}
\end{align}
\noindent The gradient of the L1 Loss can be expressed as:
\begin{equation}
\nabla_{\hat{f}}L_{\mathrm{L1}}\approx\sum_{u,v}\operatorname{sign}(F(u,v)-\hat{F}(u,v))
\end{equation}
In the frequency domain, the gradient of the L1 Loss treats both high-frequency and low-frequency components equally \cite{he2022revisiting}. However, since the spectral energy of natural images is predominantly concentrated in the low-frequency regions, the gradient of L1 Loss during optimization tends to focus more on these low-frequency components. In other words, L1 Loss prioritizes reducing low-frequency errors, which leads to smoother generated images while potentially neglecting high-frequency details. As a result, under L1 Loss optimization, high-frequency information such as edges and fine details can easily be lost, diminishing the sharpness and clarity of the image.

\noindent \textbf{Perceptual Loss.} Perceptual Loss \cite{johnson2016perceptual} is a loss function calculated in the feature space of a convolutional neural network, typically using a pre-trained network (such as VGG) to extract features from different layers to measure the differences between the generated and target images:
\begin{equation}
    L_\text{perceptual}=\sum_l\frac1{h_lw_lc_l}\|\phi_l(f)-\phi_l(\widehat{f})\|_2^2
\end{equation}
\noindent Here, $\phi_l$ represents the feature extraction function of the l-th layer of the network, and $h$, $w$, and $c$ denote the height, width, and number of channels of the feature map, respectively.

The gradient of the Perceptual Loss, calculated in the feature space, can be expressed as:
\begin{equation}
\nabla_{\hat{f}}L_{\text{perceptual}}=\sum_lH_l(u,v)\cdot(F_{\phi_l(f)}(u,v)-F_{\phi_l(\hat{f})}(u,v))
\end{equation}

\noindent Here, $H_l(u,v)$ represents the frequency response of the convolutional kernel in the $l$-th layer \cite{boashash2015time}. The frequency response of shallower convolutional layers is broader, thus $H_l(u,v)$ responds more strongly to high-frequency components \cite{fritsche2019frequency, selvaraju2017grad}. This heightened high-frequency response enhances Perceptual Loss’s sensitivity to edge and detail information, making it more effective at preserving high-frequency components during the generation process \cite{zhu2024perceptual}.

Based on the above analysis, we can conclude that L1 Loss and Perceptual Loss have different optimization directions when training a network, leading to Interest Entanglement. This implies that fine-tuning a network trained with L1 Loss using Perceptual Loss, or training a network with both losses simultaneously, will inevitably result in competition and conflict.

\subsection{Shared-Feature-Representation based Super-Resolution framework}

\noindent \textbf{SFR.} Based on previous findings, when a single model is supervised by multiple losses simultaneously or sequentially, the features learned by the model often exhibit confusion or conflict. To address this, we propose the Shared-Feature-Representation based framework (SFR). The core design concept is to have different parts of the model receive distinct supervision signals from different losses, thereby mitigating the Interest Entanglement problem through the decoupling of these supervision signals. As illustrated in the Figure \ref{fig:framework}, SFR consists of three structurally identical branches: the Regression Branch, the Perception Branch, and the Share Branch.

\begin{align}
    F_{Regression}&=E_{Regression}(I_{LR}) \\
    F_{Perceptual}&=E_{Perception}(I_{LR}) \\
    F_{Share}&=E_{Share}(I_{LR})
\end{align}
\noindent where $F$ represents the features extracted by different branches, and $E$ denotes feature extraction. We aim for the Regression Branch and Perception Branch to learn regression-related and perception-related features exclusively, focusing on high-frequency and low-frequency information of the image, respectively. Specifically, the Regression Branch receives supervision solely from the regression loss, while the Perception Branch is guided only by the perception loss. The Share Branch, however, is supervised by both types of loss, enabling it to learn beneficial, shared features from both losses.

This design allows the Share Branch to capture common, advantageous features from both objectives. Only during upsampling stage does the features are merged, thus avoiding Interest Entanglement during feature extraction. Each branch comprises three components—Shallow Extraction, Deep Extraction, and Upsample—a typical setup in image super-resolution. The Share Branch, focusing only on feature extraction, omits the Upsample component.

\noindent \textbf{InfoSqueeze.} In our experiments, we observed that directly merging features from the two branches does not effectively leverage the distinct information learned by each branch. This is due to the inherent differences in the latent spaces learned by different loss functions. To address this, we designed the InfoSqueeze module, which transforms the features learned by the Share Branch, making it easier for the other branches to integrate these features.

Since the Share Branch learns features from both types of losses, not all information in $F_{Share}$ is beneficial to the other branches. Thus, we aim for the InfoSqueeze module to filter out redundant, non-beneficial information during feature transformation. For $F_{Share} \in \mathbb{R}^{C_{{in}} \times H \times W}$, where $C_{{in}}$ is the input channel number, and $H$ and $W$ represent the height and width of the feature map, InfoSqueeze can be represented as:

\begin{equation}
    InfoSqueeze(F_{Share})=W_B*(W_A*F_{Share})
\end{equation}

\noindent Here, $W_A \in \mathbb{R}^{R \times C_{\text{in}} \times k \times k}$ and $W_B \in \mathbb{R}^{C_{\text{out}} \times R \times k \times k}$ are convolution kernels, where $k$ is the kernel size, and $R$ represents the number of channels in the intermediate layer, controlling the compression level. $W_A$ and $W_B$ serve as dimensionality reduction and expansion operators, respectively. Through this compress-and-restore approach, InfoSqueeze extracts essential features while filtering out redundant information, achieving alignment in the latent space.

The output of the Regression Branch is then:
\begin{equation}
    I_{SR}=Upsample(F_{Regression}+InfoSqueeze(F_{Share}))
\end{equation}

% ---- ---- ---- ---- ----
\begin{table*}[t]
\centering
\caption{Quantitive results on DIV2K, BSDS100, Urban100, T91, and DPED datasets for scaling factor $\times 4$. Higher PSNR and SSIM values indicate better fidelity, while lower LPIPS scores signify better perceptual quality. \textbf{Bold indicates the best performance.}}  
\scalebox{0.71}{
\begin{tabular}{ccccccccccc
>{\columncolor[HTML]{E2EFDA}}c }
\toprule
\multicolumn{2}{c}{\diagbox{Datasets}{Method}}                & DAN   & DCLS           & DASR  & MANet & SwinIR & HAT   & RGT   & RealESRGAN & ResShift      & SFR            \\ \midrule
                                  & PSNR  & 22.17 & 22.41          & 21.45 & 18.95 & 22.08  & 22.01 & 22.07 & 22.23      & 22.38         & \textbf{23.95} \\ 
                                  & SSIM  & 0.72  & 0.74           & 0.67  & 0.60  & 0.73   & 0.72  & 0.72  & 0.73       & 0.72          & \textbf{0.79}  \\
\multirow{-3}{*}{DIV2K \cite{Agustsson_2017_CVPR_Workshops}}           & LPIPS & 0.42  & 0.41           & 0.52  & 0.44  & 0.32   & 0.41  & 0.41  & 0.32       & \textbf{0.31}          & {0.32}  \\ \midrule
                                  & PSNR  & 27.95 & 28.03          & 28.79 & 22.79 & 26.08  & 28.04 & 28.01 & 26.62      & 26.40         & \textbf{29.54} \\
                                  & SSIM  & 0.87  & 0.89           & 0.89  & 0.77  & 0.84   & 0.87  & 0.87  & 0.85       & 0.81          & \textbf{0.92}  \\
\multirow{-3}{*}{BSDS100 \cite{amfm_pami2011}}         & LPIPS & 0.30  & 0.27           & 0.28  & 0.31  & 0.31   & 0.29  & 0.29  & 0.29       & 0.38          & \textbf{0.25}  \\ \midrule
                                  & PSNR  & 20.26 & 21.18          & 21.36 & 17.05 & 20.37  & 19.56 & 20.23 & 20.61      & 21.71         & \textbf{22.73} \\
                                  & SSIM  & 0.69  & 0.72           & 0.73  & 0.54  & 0.71   & 0.67  & 0.69  & 0.72       & 0.74          & \textbf{0.79}  \\
\multirow{-3}{*}{Urban100 \cite{Huang_CVPR_2015}}        & LPIPS & 0.40  & 0.37           & 0.38  & 0.40  & 0.29   & 0.39  & 0.37  & 0.29       & 0.27          & \textbf{0.27}  \\ \midrule
                                  & PSNR  & 33.14 & \textbf{33.82} & 33.64 & 27.24 & 29.41  & 33.49 & 33.49 & 29.82      & 27.93         & 32.93          \\
                                  & SSIM  & 0.93  & 0.93           & 0.94  & 0.90  & 0.92   & 0.93  & 0.93  & 0.93       & 0.85          & \textbf{0.95}  \\
\multirow{-3}{*}{T91 \cite{T91_Jianchao}}             & LPIPS & 0.23  & 0.21           & 0.21  & 0.24  & 0.28   & 0.23  & 0.22  & 0.27       & 0.40          & \textbf{0.21}  \\ \midrule
                                  & PSNR  & 22.96 & 23.38          & 23.54 & 20.22 & 22.04  & 22.75 & 22.80 & 21.89      & 22.49         & \textbf{23.95} \\
                                  & SSIM  & 0.74  & 0.76           & 0.76  & 0.64  & 0.72   & 0.74  & 0.74  & 0.72       & 0.71          & \textbf{0.78}  \\
\multirow{-3}{*}{DPED-blackberry \cite{ignatov2017dslr}} & LPIPS & 0.46  & 0.45           & 0.46  & 0.48  & 0.33   & 0.45  & 0.45  & 0.33       & 0.32          & \textbf{0.32}  \\ \midrule
                                  & PSNR  & 25.52 & 26.05          & 26.00 & 21.08 & 23.79  & 25.43 & 25.41 & 23.71      & 24.37         & \textbf{26.30} \\
                                  & SSIM  & 0.82  & 0.84           & 0.84  & 0.71  & 0.80   & 0.82  & 0.82  & 0.79       & 0.79          & \textbf{0.86}  \\
\multirow{-3}{*}{DPED-iphone \cite{ignatov2017dslr}}     & LPIPS & 0.44  & 0.42           & 0.43  & 0.46  & 0.33   & 0.42  & 0.43  & 0.33       & 0.32          & \textbf{0.31}  \\ \midrule
                                  & PSNR  & 20.43 & 20.90          & 21.02 & 18.95 & 20.72  & 20.20 & 20.24 & 20.56      & 20.86         & \textbf{21.87} \\
                                  & SSIM  & 0.64  & 0.66           & 0.66  & 0.57  & 0.65   & 0.64  & 0.64  & 0.65       & 0.63          & \textbf{0.70}  \\ 
\multirow{-3}{*}{DPED-sony \cite{ignatov2017dslr}}       & LPIPS & 0.52  & 0.51           & 0.52  & 0.53  & 0.40   & 0.52  & 0.52  & 0.39       & \textbf{0.37} & 0.41          \\ \bottomrule
\bottomrule
\end{tabular}
}
\label{tab:main}
\end{table*}
% ---- ---- ---- ---- ----

\section{Experiment}
\subsection{Experiment Setup}
\noindent \textbf{Implementation details.} The Shallow Extraction is a $3 \times 3$ convolution layer. In the Deep Extraction, we adopted RG Groups from \cite{chen2024recursive}, with the channel dimension, number of attention heads, and MLP expansion ratio set to 180, 6, and 2, respectively. $R$ in InfoSqueeze is set to 4. The Upsample component consists of a $3 \times 3$ convolution layer, a Pixel Shuffle layer, and another $3 \times 3$ convolution layer, which is a common upsampling method. During training, Regression Loss and Perceptual Loss were selected as L1 Loss and Perceptual Loss, respectively, as this combination yielded the best results. The training data generation process follows the workflow proposed in Real-ESRGAN \cite{wang2021real}. Training was conducted on four Nvidia A100 GPUs with PyTorch.

\subsection{Comparison With Existing Methods}

\noindent \textbf{Quantitative Comparisons.}
The experimental results in the Table \ref{tab:main} highlight SFR’s performance advantages across various metrics. For example, in PSNR, SFR achieved significant improvements on multiple datasets. On the DIV2K dataset, SFR reached a PSNR of 23.95, surpassing the second-best method, ResShift, by 1.57 dB (from 22.38). On the BSDS100 dataset, SFR recorded a PSNR of 29.54, outperforming DASR’s 28.79 with an increase of 0.75 dB. In the Urban100 dataset, SFR achieved a PSNR of 22.73, a 1.02 dB improvement over ResShift’s 21.71. Additionally, on the DPED-iPhone dataset, SFR’s PSNR reached 26.30, noticeably higher than DCLS’s 26.05.

In terms of SSIM, SFR also demonstrated superior structural preservation. For instance, on the BSDS100 dataset, SFR achieved an SSIM of 0.92, exceeding DCLS’s 0.89. On the Urban100 dataset, SFR recorded an SSIM of 0.79, significantly higher than ResShift’s 0.74. Moreover, on the DPED-iPhone dataset, SFR’s SSIM score was 0.86, slightly higher than DCLS’s 0.84.

In perceptual quality, as measured by the LPIPS metric, SFR also showed strong competitiveness. On the BSDS100 dataset, SFR’s LPIPS value was 0.25, better than other methods like DCLS’s 0.27. On the Urban100 dataset, SFR’s LPIPS was 0.27, comparable to the best score from ResShift. Meanwhile, on the DPED-iPhone dataset, SFR achieved an LPIPS of 0.31, a significant improvement over DCLS’s 0.42.

Overall, SFR demonstrates remarkable performance gains across multiple datasets and metrics, with particularly outstanding results in PSNR and SSIM, showcasing its excellent capabilities in restoring image details and preserving structural integrity. Its LPIPS performance further confirms the consistency and high quality of the generated images in terms of perceptual alignment, indicating that SFR effectively balances fidelity and perceptual quality.

% ---- figure ----  
\begin{figure*}[t]
\centering
 \includegraphics[width=\textwidth]{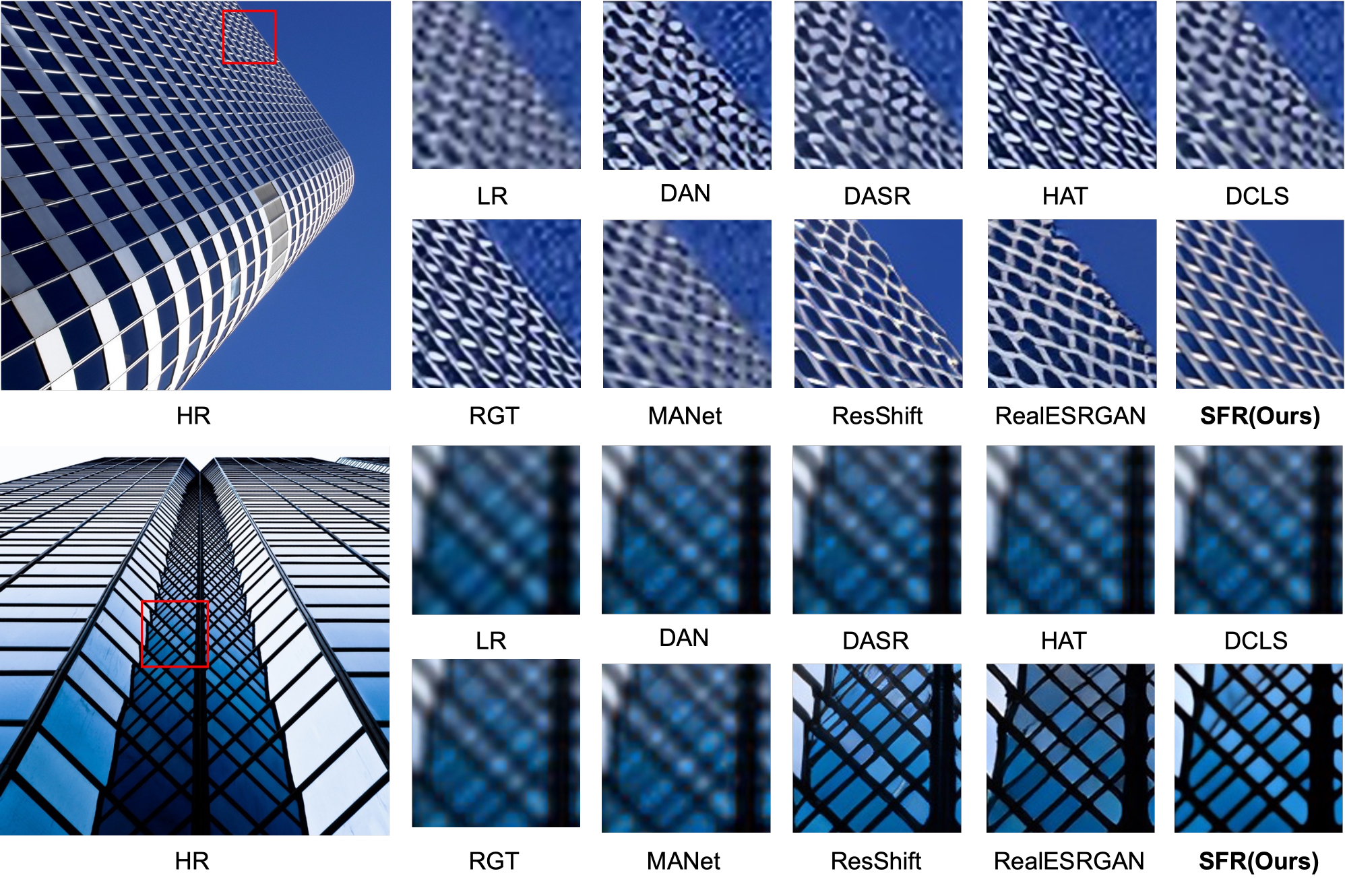}
 \caption{Visual comparisons of several representative methods on examples of the Urban100 dataset.}
   \label{fig:urban}
\end{figure*}
% ---- ---- ---- ---- ----
\noindent \textbf{Qualitative Comparisons.} In Figure \ref{fig:urban}, due to the inherent challenges of the blind SR task, it is often difficult for SR models to strike a balance between fidelity and perceptual quality; methods that excel in pixel-based metrics do not necessarily deliver satisfactory visual results. Compared with other methods, SFR not only leads in quantitative metrics but also excels in restoring image details.

We observe a comparison of SR reconstruction effects for two urban buildings. Initially, the high-resolution image provides exceptionally sharp and detailed architectural features, including the clear edges of windows and detailed textures of glass reflections. Conversely, the low-resolution image exhibits significant loss of detail, with the building's lines and patterns becoming blurred.

The magnified area highlights the performance differences among various methods in SR reconstruction. In the comparison of buildings, the SFR (Ours) method demonstrates particularly effective restoration of details and patterns, with the lines between windows clearly visible and the patterns closely resembling those of the high-resolution original image. Moreover, the SFR method successfully restores the geometric symmetry of the buildings, a challenge for other methods which sometimes introduce blurring or ripple-like artifacts at these edge areas.

When compared with other methods such as RGT \cite{chen2024recursive}, MANet \cite{liang2021mutual}, ResShift \cite{yue2024resshift}, and RealESRGAN \cite{wang2021real}, it is evident that they face varying degrees of challenges when dealing with such complex structures. Some methods may perform well in restoring details but fall short in maintaining straight lines and natural textures. In contrast, the SFR method not only delivers high-quality details but also exhibits significant advantages in overall geometric fidelity and visual impact of the images.

In summary, SFR provides a more natural and smoother overall visual experience, maintaining high fidelity across all aspects, from subtle textures to larger structures. Its performance surpasses other SR methods across multiple dimensions, including detail sharpening, color accuracy, and avoiding overly processed artifacts.

\begin{table}[t]
\centering
\caption{Ablation study on different branches on DIV2K dataset.}  
\scalebox{0.85}{
\begin{tabular}{cccccc}
\toprule
\begin{tabular}[c]{@{}c@{}}Regression \\ Branch\end{tabular} & \begin{tabular}[c]{@{}c@{}}Share \\ Branch\end{tabular} & \begin{tabular}[c]{@{}c@{}}Perception \\ Branch\end{tabular} & PSNR           & SSIM          & LPIPS         \\ \midrule
\cmark                 & \xmark          & \xmark               & 20.82          & 0.68          & 0.34          \\ 
\cmark                 & \cmark          & \xmark               & 19.62          & 0.58          & 0.46          \\
\cmark                 & \xmark          & \cmark               & 22.48          & 0.74          & 0.45          \\
\rowcolor[HTML]{E2EFDA} 
\cmark                 & \cmark          & \cmark               & \textbf{23.95} & \textbf{0.79} & \textbf{0.32} \\ \bottomrule \bottomrule
\end{tabular}}
\label{tab:branch}
\end{table}
% ---- ---- ---- ---- ----

\subsection{Ablation Study}
\noindent \textbf{Ablation on branches.} To demonstrate the effectiveness of SFR’s multi-branch design, we conducted ablation experiments on the network branches. Table \ref{tab:branch} shows the impact of different branch combinations on model performance. The complete model (last row), achieves the best performance across all metrics, with PSNR, SSIM, and LPIPS reaching 23.95, 0.79, and 0.32, respectively, indicating an excellent balance between image quality and perceptual effect.

When the Share Branch is removed (first row), PSNR and SSIM drop to 20.82 and 0.68, while LPIPS rises to 0.34, showing a decline in quality, especially with imbalanced structural and perceptual consistency. Removing the Perceptual Branch (second row) leads to a further decrease in PSNR and SSIM to 19.62 and 0.58, and LPIPS increases to 0.46, suggesting that simply adding branches does not necessarily enhance performance. When only the Regression and Perceptual Branches are retained (third row), PSNR and SSIM remain relatively high, but LPIPS rises to 0.45, highlighting the Share Branch’s crucial role in balancing image quality and perceptual consistency.

\begin{figure}[t]
  \centering
  % \fbox{\rule{0pt}{2in} \rule{0.97\linewidth}{0pt}}
    \includegraphics[width=0.89\linewidth]{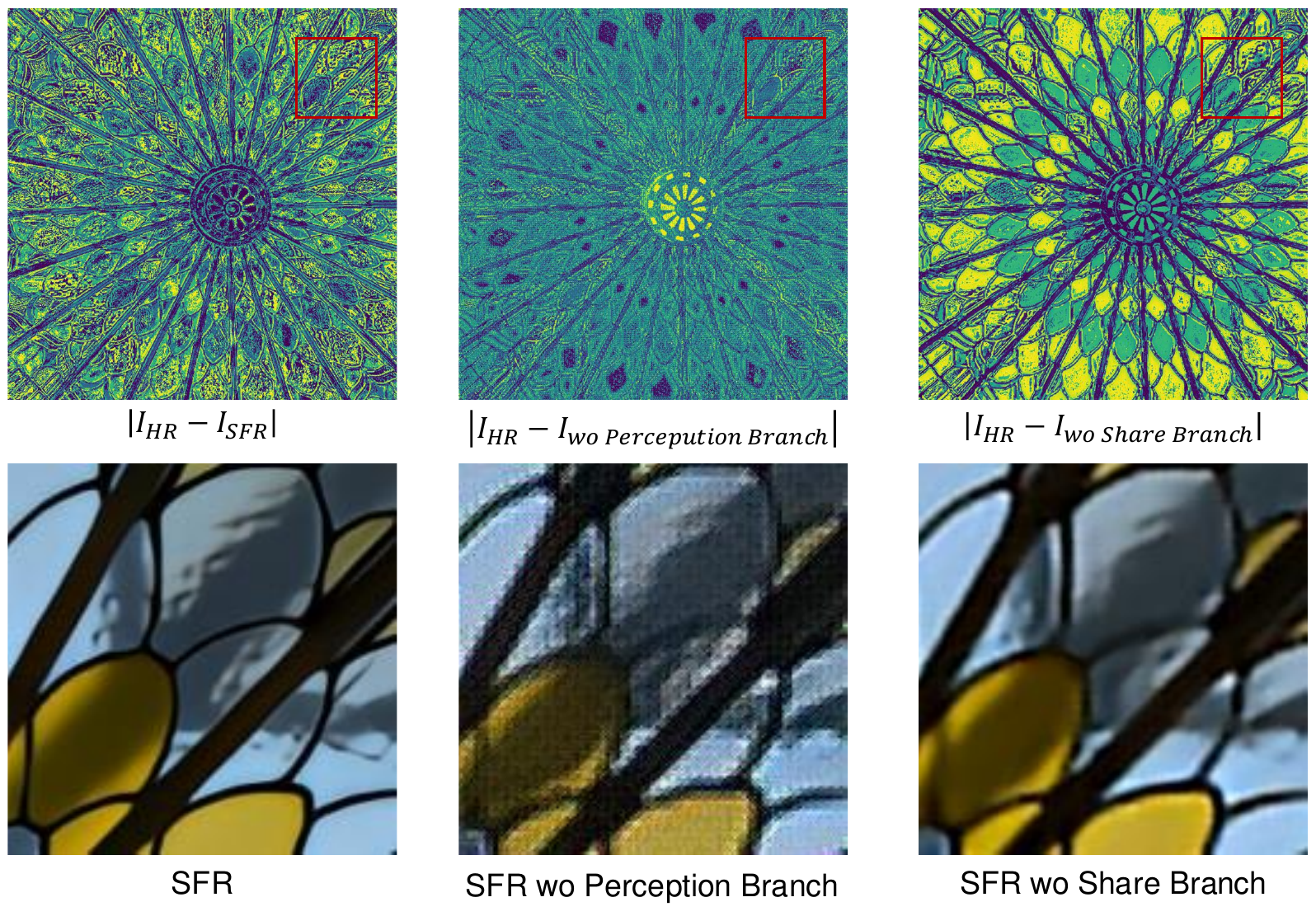}
   \caption{Visualization of ablation study on different branches. The first row shows the residuals between the HR and different images, while the second row displays the corresponding magnified original images.}
   \label{fig:minus}
\end{figure}

Figure \ref{fig:minus} illustrates the residuals between the HR image and different branch combinations, as well as localized super-resolved image displays. From Figure \ref{fig:minus}, we observe that, in the absence of the Perception Branch, although the absolute difference from the original image is reduced, the resulting image exhibits noticeable noise and blurring. When the Share Branch, which acts as an intermediary buffer, is missing, the residual image shows large patches of localized differences, reflected as excessive smoothing in the SR image. This demonstrates that the SFR design effectively balances fidelity and perceptual quality.

Overall, each branch significantly contributes to model performance, and removing any branch results in imbalances across metrics, underscoring the importance of the complete branch design.
% ---- figure ----  
% ---- ---- ---- ---- ----

\begin{table}[h]
\centering
\caption{Comparison of different feature transform methods on DIV2K dataset.}  
\scalebox{0.78}{
\begin{tabular}{cccc}
\toprule
Feature   Transform Method & PSNR           & SSIM          & LPIPS         \\ \midrule
ADD                 & 23.65          & 0.78          & 0.39          \\
SFT                 & 23.46          & 0.77          & 0.41          \\
\rowcolor[HTML]{E2EFDA} 
InfoSqueeze (Ours)   & \textbf{23.95} & \textbf{0.79} & \textbf{0.32} \\ \bottomrule \bottomrule
\end{tabular}}
\label{tab:transfrom}
\end{table}
% ---- ---- ---- ---- ----
\noindent \textbf{Importance of InfoSqueeze.} As discussed previously, due to the distinct feature spaces of each branch, a simple feature interaction approach cannot fully harness SFR’s potential. Table \ref{tab:transfrom} shows the impact of different feature transformation methods on model performance, highlighting the advantages of InfoSqueeze. ADD represents directly adding the features from the two branches, while Spatial Feature Transform (SFT) \cite{wang2018recovering} is a widely used feature fusion method in the field of image SR. When using InfoSqueeze for feature transformation, the model achieves optimal performance across PSNR, SSIM, and LPIPS, with scores of 23.95, 0.79, and 0.32, respectively, indicating a significant improvement in both image reconstruction quality and perceptual consistency.

In contrast, traditional feature transformation methods like ADD and SFT demonstrate some limitations in balancing different losses. With ADD, the model’s PSNR and SSIM reach 23.65 and 0.78, while LPIPS rises to 0.39, indicating a slight decrease in perceptual quality. SFT performs slightly lower than ADD, with PSNR at 23.46, SSIM at 0.77, and LPIPS at 0.41, showing that it is less effective than InfoSqueeze in balancing losses.

In summary, InfoSqueeze effectively unlocks SFR’s potential, significantly enhancing perceptual quality while maintaining numerical reconstruction accuracy, thereby achieving a better balance between different losses.

\noindent \textbf{Comparison on Different Losses.} Table \ref{tab:loss} presents the impact of different combinations of regression losses (L1 and MSE) and perceptual losses (GAN and Perceptual) on model performance. The combination of L1 and Perceptual loss achieves the best results, with PSNR, SSIM, and LPIPS reaching 23.95, 0.79, and 0.32, respectively.

In comparison, other loss combinations perform slightly less effectively. For example, the L1 and GAN loss combination achieves a PSNR of 23.44 and an SSIM of 0.77, but LPIPS increases to 0.42, indicating a slight decline in perceptual quality. The MSE and GAN combination yields a slightly higher PSNR of 23.94, but with a higher LPIPS of 0.39, showing minor shortcomings in perceptual consistency. When using MSE with Perceptual loss (23.58 PSNR and 0.77 SSIM), LPIPS reaches 0.37; although perceptual quality improves slightly, overall performance does not match that of the L1 and Perceptual combination.

\begin{table}[h]
\centering
\caption{Comparison of losses combinations on DIV2K dataset.}
\scalebox{0.76}{
\begin{tabular}{ccccc}
\toprule
Regression   Loss & Perception Loss          & PSNR           & SSIM          & LPIPS         \\ \midrule
L1                & GAN                      & 23.44          & 0.77          & 0.42          \\
MSE               & GAN                      & 23.94          & 0.78          & 0.39          \\
\rowcolor[HTML]{E2EFDA} 
L1                & Perceptual & \textbf{23.95} & \textbf{0.79} & \textbf{0.32} \\
MSE               & Perceptual          & 23.58          & 0.77          & 0.37          \\ \bottomrule \bottomrule
\end{tabular}}
\label{tab:loss}
\end{table}
% ---- ---- ---- ---- ----

\section{Conclusion}
In summary, this paper analyzes the Interest Entanglement issue in blind image super-resolution from a frequency domain perspective and proposes the Shared-Feature-Representation based Super-Resolution framework (SFR) accordingly. By decoupling the learning processes of different losses within the network, SFR effectively balances the conflicting and competing optimization objectives of fidelity and perceptual quality. In future work, we plan to explore the application of the SFR framework to other low-level vision tasks.

\bibliography{aaai2026}


\end{document}